# EFFECTIVE FAQ RETRIEVAL AND QUESTION MATCHING WITH UNSUPERVISED KNOWLEDGE INJECTION

*Wen-Ting Tseng†, Tien-Hong Lo†, Yung-Chang Hsu† and Berlin Chen†*

†National Taiwan Normal University, Taiwan
†EZAI, Taiwan

## ABSTRACT

Frequently asked question (FAQ) retrieval, with the purpose of providing information on frequent questions or concerns, has far-reaching applications in many areas like e-commerce services, online forums and many others, where a collection of question-answer (*Q-A*) pairs compiled a priori can be employed to retrieve an appropriate answer in response to a user's query that is likely to reoccur frequently. To this end, predominant approaches to FAQ retrieval typically rank question-answer pairs by considering either the similarity between the query and a question (*q-Q*), the relevance between the query and the associated answer of a question (*q-A*), or combining the clues gathered from the *q-Q* similarity measure and the *q-A* relevance measure. In this paper, we extend this line of research by combining the clues gathered from the *q-Q* similarity measure and the *q-A* relevance measure, and meanwhile injecting extra word interaction information, distilled from a generic (open-domain) knowledge base, into a contextual language model for inferring the *q-A* relevance. Furthermore, we also explore to capitalize on domain-specific topically-relevant relations between words in an unsupervised manner, acting as a surrogate to the supervised domain-specific knowledge base information. As such, it enables the model to equip sentence representations with the knowledge about domain-specific and topically-relevant relations among words, thereby providing a better *q-A* relevance measure. We evaluate variants of our approach on a publicly-available Chinese FAQ dataset (viz. TaipeiQA), and further apply and contextualize it to a large-scale question-matching task (viz. LCQMC), which aims to search questions from a QA dataset that have a similar intent as an input query. Extensive experimental results on these two datasets confirm the promising performance of the proposed approach in relation to some state-of-the-art ones.

*Index Terms*— FAQ, question-matching, knowledge graph, language model

## 1. INTRODUCTION

With the explosive growth of text (and multimedia) information repositories and services on the Internet, developments of effective frequently asked question (FAQ) retrieval [1] techniques have become a pressing need for a wide variety of application areas, like e-commerce services, online forums and so forth. FAQ retrieval has the goal of providing information on frequent questions or concerns, which is fulfilled by leveraging a collection of question-answer (denoted by *Q-A*) pairs compiled ahead of time to search an appropriate answer in response to a user's query (denoted by *q* for short) that is supposed to reoccur frequently. Recently, a common thread in various approaches to FAQ retrieval has been to rank question-answer pairs by considering either the similarity between the query and a question (viz. the *q-Q* similarity measure), or the relevance between the query and the associated answer of a question (viz. the q-A relevance measure). For example, the *q-Q* similarity measure can be computed with unsupervised information retrieval (IR) models, such as the Okapi BM25 method [2] and the vector space method [3], to name a few. Meanwhile, the *q-A* relevance can be determined with a simple supervised neural model stacked on top of a pre-trained contextual language model, which takes a query as the input and predicts the likelihoods of all answers given the query. Prevailing contextual language models, such as the bidirectional encoder representations from transformers (BERT) [4], embeddings from language models (ELMo) [5], generative pre-trained transformer (GPT), and the generalized autoregressive pretraining method (XLNet) [6], can serve this purpose to obtain context-aware query embeddings. Among them, BERT has recently aroused much attention due to its excellent performance on capturing semantic interactions between two or more text units.

The supervised neural model usually is trained (and the contextual language model is fine-tuned) on the *Q-A* pairs collected a priori for a given FAQ retrieval task [7]. Although such pre-trained contextual language models can learn general language representations from large-scale corpora, they may fail to capture the important open-domain or domain-specific knowledge about deeper semantic and pragmatic interactions of entities (or words) involved in a given knowledge-driven FAQ retrieval task. As such, there is good reason to explore extra open-domain or domain-specific knowledge clues for use in FAQ retrieval. However, manually building a domain-specific knowledge base could be tedious and expensive in terms of time and personnel.

Building on these insights, we propose an effective approach to FAQ retrieval, which has at least three distinctive characteristics. First, both the *q-Q* similarity measure obtained by the Okapi BM25 method and the *q-A* relevance measure obtained by a BERT-based supervised method are linearly combined to rank *Q-A* pairs for better retrieval performance. On one hand, Okapi BM25 performs bag-of-word term matching between the query and a question, it can facilitate high-precision retrieval. On the other hand, since the BERT-based supervised method determines the relevance between the query and an answer based on context-aware semantic embeddings, which can model long-range dependency and get around the term-mismatch problem to some extent. In addition, an effective voting mechanism to rerank answer hypotheses for better performance is proposed. Second, inspired by the notion of knowledge-enabled BERT (K-BERT) modeling [8], we investigate to inject triplets of entity relations distilled from an open-domain knowledge base into BERT to expand and refine the representations of an input query for more accurate relevance estimation. Third, since a domain-specific knowledge base is not always readily available, we leverage an unsupervised topic modeling method, viz. probabilistic latent topic analysis (PLSA) [9], to extract triplets of topically-relevant words from the collection of Q-A pairs, which can enrich the query representations for the FAQ retrieval task at hand. To further confirm the utility of our approach, we further apply and contextualize it to question-matching, which is highly relevant to FAQ retrieval and has drawn increasing attention recently.

## 2. APPROACH TO FAQ RETRIEVAL AND QUESTION MATCHING

This section first sheds light on the instantiations of the two disparate measures that we employ to rank the collection of Q-A pairs given that a user's query is posed for FAQ retrieval: 1) the $q$-$Q$ similarity measure and 2) the $q$-$A$ relevance measure. An effective voting mechanism is proposed to leverage the clues gathered from the $q$-$Q$ similarity measure and the $q$-$A$ relevance measure for selecting a more accurate answer. After that, we introduce the ways that the triplets of entity-level or word-level semantic and pragmatic relations extracted from an open-domain knowledge base and in-domain topical clusters are incorporated into the BERT-based method to increase the accuracy of the $q$-$A$ relevance measure.

### 2.1. The $q$-$Q$ Similarity Measure

The task of FAQ retrieval is to rank a collection of question-answer (Q-A) pairs, $\{(Q_1, A_1), \ldots, (Q_n, A_n), \ldots, (Q_N, A_N)\}$, with respect to a user's query, and then return the answer of the topmost ranked one as the desired answer. Note here that a distinct answer may be associated with different questions, which means that the number of distinct answers may be smaller than or equal to $N$. Furthermore, due to the fact that a query may not have been seen before, a common first thought is to calculate the $q$-$Q$ similarity measure by resorting to an unsupervised method developed by the IR community, such as the Okapi BM25 method [2]:

$$BM25(q, Q_n) = \sum_{l=1}^{L} \frac{(K_1 + 1)f(w_l, Q_n)}{K_1 \left[(1-b) + b \frac{len(Q_n)}{avg_{Qlen}}\right] + f(w_l, Q_n)} IQF(w_l), \quad (1)$$

where the query $q$ can be expressed by $q = w_1, \ldots, w_L$, $f(w_l, Q_n)$ is the frequency of word $w_l$ within the question $Q_n$, $len(Q_n)$ is the length of $Q_n$, $avg\_Qlen$ denotes the average question length for the collection, and $IQF(w_l)$ is the inverse question frequency of $w_l$. In addition, $K_1$ and $b$ are tunable parameters.

### 2.2. The $q$-$A$ Relevance Measure

Apart from calculating the $q$-$Q$ similarity measure, we can also estimate the $q$-$A$ relevance measure for ranking the collection of Q-A pairs. Instead of using the unsupervised Okapi BM25 method, we in this paper employ a supervised neural method for this purpose, which encompasses a single layer neural network stacked on top of a pre-trained BERT-based neural network. In the test phase, the whole model will accept an arbitrary query $q$ as the input and its output layer will predict the posterior probability $P(A_n|q)$, $n = 1, \ldots, N$, of any answer $A_n$ (denoted also by $BERT(q, A_n)$). The answer $A_n$ that has the highest $P(A_n|q)$ value will be regarded as the desired answer that is anticipated to be the most relevant to $q$. On the other side, in the training phase, since the test queries are not given in advance, we can instead capitalize on the corresponding relations of existing Q-A pairs for model training. More specifically, the one-layer neural network is trained (and meanwhile the parameters of BERT are fine-tuned) by maximizing the $P(A_n|Q_n)$ for all the Q-A pairs in the collection. Following a similar vein, other supervised methods based on contextual language models, such as BiLSTM, CNN, ELMO, BiMPM, ELMo, and many others, can also be used to calculate $P(A_n|q)$ [5], [7], [10].

To bring together the modeling advantages of the $q$-$Q$ similarity measure and the $q$-$A$ relevance measure, the ultimate ranking score $RS(Q_n, A_n, q)$ for a Q-A pair with respect to a query $q$ can be obtained through the following linear combination:

$$RS(Q_n, A_n, q) = \alpha \cdot \frac{BM25(q, Q_n)}{\sum_{n'=1}^{N} BM25(q, Q_{n'})} + (1-\alpha) \cdot BERT(q, A_n), \quad (2)$$

where $\alpha$ is tunable parameter used to control the relative contributions of the two measures.

### 2.3. Voting Mechanism

Apart from simply selecting the answer that corresponds to Q-A pair that has the highest $RS(Q_n, A_n, q)$ score, we can instead leverage a voting mechanism to pick up the answer that has the majority occurrence count in the top-$M$ ranked results in terms of the $RS(Q_n, A_n, q)$ scores for all the Q-A pairs (in our experiments, we set $M$ to 5). Namely, if there is an answer that has an occurrence count greater or equal to $[M/2]$ in the top-$M$ ranked results, it will be selected as the target answer of the input query $q$. Otherwise, the answer corresponding to the highest $RS(Q_n, A_n, q)$ score will be selected in response to $q$.

### 2.4. Supervised Knowledge Injection for the $q$-$A$ Relevance Measure

The aforementioned BERT-based method for estimating the $q$-$A$ relevance measure employs a pre-trained Transformer [11] architecture, equipped with multi-head attention mechanisms, which has been proven effective for capturing semantic interactions between text units [4]. However, it might not perform well on knowledge-driven tasks like FAQ retrieval or question-matching, due to the incapability of modelling open-domain or/and domain-specific knowledge about deeper semantic and pragmatic interactions of words (entities). To ameliorate this problem, a new wave of research has emerged recently to incorporate information gleaned from an open-domain knowledge base [12], such as WordNet [13], HowNet [14], YAGO [15], or a domain-specific knowledge base, such as MedicalKG, into the BERT-based model structure. Representative methods include, but is not limited to, the THU-ERNE [16] method and the Knowledge-enabled BERT (K-BERT) method. On the practical side, K-BERT seems more attractive than THU-ERNE, because it can easily inject a given open-domain or domain-specific knowledge base, in the form of a set of triplets $(w_i, \text{relation}_r, w_j)$ that describe disparate relations between words or entities, into a pretrained BERT-based model structure through the so-called soft-position and visible-matrix operations [8]. As such, in this paper, we make a further attempt to exploit K-BERT to incorporate both open-domain and domain-specific knowledge clues for use in the *q-A* relevance measure [17], [18], [19], [20], [21].

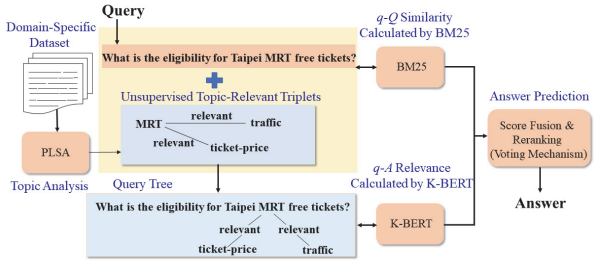

Figure 1: A schematic depiction of our proposed modelling approach to unsupervised knowledge injection.

| Topic 1 | Topic 2 | … | Topic 10 |
|---------|---------|---|----------|
| Taipei  | library |   | Loan     |
| city    | book    |   | Pioneer  |
| MRT     | read    | . | Young    |
| Taiwan  | information | . | Financing |
| capital | data    | . | Bank     |
| .       | .       |   | .        |
| .       | .       |   | .        |
| .       | .       |   | .        |

Table 1: An illustration of the words that have the largest probability values of $P(w|T_k)$ for some topics of the TaipeiQA task.

### 2.5. Unsupervised Knowledge Injection for the *q-A* Relevance Measure

Since hand-crafting a domain-specific knowledge base for an FAQ retrieval task (or a question-matching task) would be extremely laborious and time-consuming, in this paper we alternatively employ probabilistic latent topic analysis (PLSA), a celebrated unsupervised topic modelling method, to extract the domain-specific word-level topical relations, incapsulated in the collection of *Q-A* pairs for the FAQ retrieval task of interest. Given the collection of question-answer (*Q-A*) pairs, $\{(Q_1, A_1), \ldots, (Q_n, A_n), \ldots, (Q_N, A_N)\}$, the PLSA formulation decomposes the word-usage distribution of a *Q-A* pair by

$$P(w|Q_n, A_n) = \sum_{k=1}^{K} P(w|T_k) P(T_k|Q_n, A_n), \quad (3)$$

where $w$ denotes an arbitrary word, $T_k$ marks an automatically generated topic and $K$ is the total number of topics (in our experiments, we set $K$ to 10 by default). More specifically, we assume the top significant words of a given PLSA-generated topic

[1] https://github.com/DTDwind/TaipeiQA

$T_k$ have a domain-specific topically-relevant relation, which can also be represented in triplets as well to be digested by K-BERT. As such, for each topic $T_k$, we can select those top-*L* words, 10 words for example, which have the largest probability values of $P(w|T_k)$ and establish symmetric topically-relevant relations among them, namely $(w_i, \text{relevance}_{T_k}, w_j)$ and $(w_j, \text{relevance}_{T_k}, w_i)$ for any pair of words $(w_i, w_j)$ involved in the Top-*L* words. To our knowledge, this is the first attempt to harness the synergistic power of unsupervised domain-specific, topically-relevant word relations and K-BERT for use in the *q-A* relevance measure for FAQ retrieval. Along this same vein, use can capitalize on PLSA to generate domain-specific topically-relevant relations of words for the question-matching task as well. Figure 1 shows a schematic depiction of our proposed modelling approach to unsupervised knowledge injection, while Table is an illustration of the words that have the largest probability values of $P(w|T_k)$ for some topics of the TaipeiQA task.

### 3. EXPERIMENTAL DATASETS AND SETUP

We assess the effectiveness of our proposed approach on two publicly-available Chinese datasets: TaipeiQA[1] and LCQMC [10]. TaipeiQA is an FAQ dataset crawled from the official website of the Taipei City Government, which consists of 8,321 *Q-A* pairs and is further divided into three parts: the training set (68%), the validation set (20%) and the test set (12%). Note here that the questions in the validation and test sets are taken as unseen queries, which are used to tune the model parameters and evaluate the performance of FAQ retrieval, respectively. On the other hand, LCQMC is a publicly-available large-scale Chinese question-matching corpus, which consists of 260,068 question pairs and is further divided into three parts: the training set (60%), the validation set (20%) and the test set (20%). The task of LCQMC is to determine the similarity of a pair of questions in terms of their intents, which is more general than paraphrasing. In implementation for the LCQMC task, the input of Okapi BM25 is a pair of questions and the output is the corresponding similarity measure. As for the BERT-based supervised method, its input to BERT (and K-BERT) is changed to be a pair of questions, while the output of the upper-stacked one-layer neural network is a real value between 0 and 1 that quantifies the similarity between the two (the larger the value, the higher the similarity). In addition, for both the TaipeiQA and LCQMC tasks, HowNet will be taken as the open-domain knowledge base.

### 4. EXPERIMENTAL RESULTS

In the first set of experiments, we assess the performance of our baseline approach (denoted by BM25+BERT) on the TaipeiQA task with different evaluation metric [22], [23], in relation to those approaches either using unsupervised method (BM25) or supervised methods (BiLSTM and ELMO) in isolation. As can be seen from the first five rows of Table 2, BM25+BERT performs better than the other approaches for most evaluation metrics, which reveals the complementarity between Okapi BM25 and BERT for the FQA retrieval task studied here.

In the second set of experiments, we evaluate the effectiveness of the enhanced BERT modelling (viz. K-BERT), which has been injected with knowledge clues either from the open domain (viz. K-BERT(HowNet)) [24], [25], from the unsupervised domain-specific topical clusters (viz. K-BERT(TaipeiQA)), or from their combination (viz. K-BERT(HowNet+TaipeiQA)). It is evident from Table 2, K-BERT(TaipeiQA) offers considerable improvements over BERT and K-BERT(HowNet). This indeed confirms the utility of our unsupervised topical knowledge distilling method (*cf.* Sections 2.4 and 2.5). The last row of Table 3 depicts the results of the combination of Okapi BM25 and K-BERT(TaipeiQA), which yields the best results for the TaipeiQA task.

Table 2: Evaluations of various modelling approaches on the TaipeiQA task.

|  | Precision | Recall | F1 | Accuracy | MRR |
|---|---|---|---|---|---|
| BM25 | 0.743 | 0.700 | 0.692 | 0.743 | 0.775 |
| BERT | 0.688 | 0.675 | 0.651 | 0.697 | 0.771 |
| ELMO | 0.738 | 0.570 | 0.643 | 0.466 | 0.528 |
| BILSTM | 0.798 | 0.577 | 0.670 | 0.439 | 0.501 |
| BM25+BERT | 0.763 | 0.747 | 0.730 | 0.785 | 0.798 |
| K-BERT(HowNet) | 0.705 | 0.685 | 0.665 | 0.706 | 0.774 |
| K-BERT(TaipeiQA) | 0.757 | 0.724 | 0.712 | 0.731 | 0.791 |
| K-BERT (HowNet+TaipeiQA) | 0.754 | 0.722 | 0.712 | 0.726 | 0.788 |
| BM25+K-BERT (TaipeiQA) | 0.812 | 0.792 | 0.781 | 0.811 | 0.802 |
| BM25+KBERT (TaipeiQA)[VOTE] | **0.813** | **0.793** | **0.782** | **0.812** | **0.807** |

Table 3: Evaluations of various modelling approaches on the LCQMC task.

|  | Precision | Recall | F1 | Accuracy |
|---|---|---|---|---|
| BM25 | 0.659 | 0.659 | 0.658 | 0.659 |
| BERT | 0.855 | 0.842 | 0.841 | 0.842 |
| BiLSTM | 0.706 | 0.893 | 0.789 | 0.761 |
| CNN | 0.671 | 0.856 | 0.752 | 0.718 |
| BiMPM | 0.776 | 0.939 | 0.850 | 0.834 |
| BM25+BERT | 0.856 | 0.842 | 0.849 | 0.845 |
| K-BERT(HowNet) | 0.859 | 0.872 | 0.858 | 0.859 |
| K-BERT(LCQMC) | 0.877 | 0.870 | 0.870 | 0.870 |
| K-BERT (HowNet+LCQMC) | 0.873 | 0.862 | 0.861 | 0.862 |
| BM25+K-BERT (LCQMC) | 0.877 | 0.870 | 0.870 | 0.870 |
| BM25+K-BERT (LCQMC) [VOTE] | **0.877** | **0.870** | **0.870** | **0.871** |

In the third set of experiments, we report on the results of our baseline and enhanced approaches on the LCQMC question-matching task, in comparison to some unsupervised and supervised approaches, where the results of the three supervised approaches (viz. BiLSTM, CNN and BiMPM) are adopted from [10]. Similar trends can be observed from Table 3 BM25+K-BERT(LCQMC) performs remarkably well for the LCQMC task.

As a final point in this section, we evaluate the FAQ performance levels of K-BERT(TaipeiQA) with respect to different numbers of the PLSA topics used for extracting the domain-specific knowledge clues, as shown in Figure 2. Consulting Figure 2 we notice that the performance of K-BERT(TaipeiQA) is steadily improved when the number of PLSA being used becomes larger; the improvement, however, seems to reach a peak elevation when the number is set to 10, and then to degrade when the number exceeds 10. However, how to automatically determine the number of topics, as well as the number of significant words for each topic to establish domain-specific topically-relevant relations, still awaits further studies.

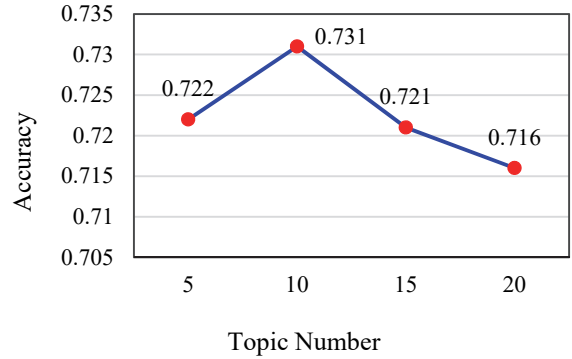

Figure 2: The FAQ performance levels of K-BERT(TaipeiQA) with respect to different numbers of the PLSA topics.

## 5. CONCLUSION AND FUTURE WORK

In this paper, we have presented an effective, hybrid approach for FAQ retrieval, exploring the synergistic effect of combing unsupervised IR method (Okapi BM25) and supervised contextual language models (BERT and K-BERT). In addition, a novel knowledge injection mechanism leveraging clues drawn from unsupervised domain-specific topical clusters has been proposed. As to future work, we plan to investigate more sophisticated unsupervised knowledge injection mechanisms and learning algorithms, as well as their applications to document summarization, readability assessment and among others.

## 6. ACKNOWLEDGMENT

This research is supported in part by ASUS AICS and the Ministry of Science and Technology (MOST), Taiwan, under Grant Number MOST 109-2634-F-008-006- through Pervasive Artificial Intelligence Research (PAIR) Labs, Taiwan, and Grant Numbers MOST 108-2221-E-003-005-MY3 and MOST 109-2221-E-003-020-MY3. Any findings and implications in the paper do not necessarily reflect those of the sponsors.

## 7. REFERENCES

[1] Mladen Karan and Jan Šnajder. 2018. Paraphrase-focused learning to rank for domain-specific frequently asked questions retrieval. *Expert Systems with Applications*, 91: 418-433.

[2] Stephen Robertson and Hugo Zaragoza. 2009. The

probabilistic relevance framework: BM25 and beyond, *Foundations and Trends in Information Retrieval,* 3(4): 333–389.

[3] Gerard Salton, Andrew Wong, and Chungshu Yang. 1975. A vector space model for automatic indexing. *Communications of the ACM, 18(11),* pages 613–620.

[4] Jacob Devlin, Ming-Wei Chang, et al. 2019. Bert: Pre-training of deep bidirectional transformers for language understanding. In *Proceedings of the Conference of the North American Chapter of the Association for Computational Linguistics: Human Language Technologies.* pages 4171–4186.

[5] Matthew Peters, Mark Neumann, et al. 2018. Deep contextualized word representations. In *Proceedings of the Conference of the North American Chapter of the Association for Computational Linguistics: Human Lan-guage Technologies,* pages 2227–2237.

[6] Zhilin Yang, Zihang Dai, et al. 2019. XLNet: Generalized autoregressive pretraining for language understanding. In *Proceedings of Conference on Neural Information Processing Systems,* pages 5753–5763.

[7] Wataru Sakata, Tomohide Shibata, et al. 2019. FAQ retrieval using query-question similarity and BERT-based query-answer relevance. In *Proceedings of the International ACM SIGIR Conference on Research and Devel-opment in Information Retrieval,* pages 1113–1116.

[8] Weijie Liu, Peng Zhou, et al. 2020. K-BERT: enabling language representation with knowledge graph. In *Proceedings of the AAAI Conference on Artificial Intelligence AAAI,* pages 2901–2908.

[9] Thomas Hofmann. 1999. Probabilistic latent semantic analysis. In Proceedings of the Conference on Uncertain-ty in Artificial Intelligence, pages 289–296.

[10] Xin Liu, Qingcai Chen, et al. 2018. LCQMC: a large-scale Chinese question matching corpus. In *Proceedings of the International Conference on Computational Linguistics*, pages 1952–1962.

[11] Ashish Vaswani, Noam Shazeer, et al. 2017. Attention is all you need. In *Proceedings of Conference on Neural Information Processing Systems*, pages 5998–6008.

[12] Wanyun Cuix, Yanghua Xiao, et al. 2017. KBQA: learning question answering over QA corpora and knowledge bases. *Proceedings of the VLDB Endowment*, 10(5): 656–676.

[13] George A. Miller. 1995. WordNet: a lexical database for English. *Communications of the ACM*, *38*(11): 39–41.

[14] Zhendong Dong, Qiang Dong, and Changling Hao. 2010. HowNet and Its Computation of Meaning. In *Proceed-ings of the International Conference on Computational Linguistics,* pages 53–56.

[15] Fabian M. Suchanek, Gjergji Kasneci, and Gerhard Weikum. 2007. YAGO: a core of semantic knowledge. In *Proceedings of the international conference on World Wide Web,* pages 697–706.

[16] Zhengyan Zhang, Xu Han, et al. 2019. ERNIE: enhanced language representation with informative entities, In *Proceedings of the Annual Meeting of the Association for Computational Linguistics,* pages 1441–1451.

[17] Liang Yao, Chengsheng Mao, and Yuan Luo. 2019. KG-BERT: BERT for knowledge graph completion. arXiv preprint arXiv:1909.03193.

[18] Shaoxiong Ji, Shirui Pan, et al. 2020. A survey on knowledge graphs: Representation, acquisition and applications. arXiv preprint arXiv:2002.00388.

[19] Quan Wang, Mao Zhendong, et al. 2017. Knowledge graph embedding: A survey of approaches and applications. *IEEE Transactions on Knowledge and Data Engineering*, *29*(12), 2724-2743.

[20] Guoliang Ji, Shizhu He, et al. 2015. Knowledge graph embedding via dynamic mapping matrix. In *Proceedings of the 53rd annual meeting of the association for computational linguistics and the 7th international joint conference on natural language processing (volume 1: Long papers)*, pages 687-696.

[21] Guoliang Ji, Kang Liu, et al. 2016. Knowledge graph completion with adaptive sparse transfer matrix. In *Thirtieth AAAI conference on artificial intelligence*.

[22] Marina Sokolova and Guy Lapalme, 2009. A systematic analysis of performance measures for classification tasks. *Information processing & management*, *45*(4): 427–437.

[23] Svetlana Kiritchenko, Matwin Stan, et al. 2006. Learning and evaluation in the presence of class hierarchies: Application to text categorization. In *Conference of the Canadian Society for Computational Studies of Intelligence,* pages 395-406.

[24] Zhen Wang, Jianwen Zhang, et al. 2014. Knowledge graph and text jointly embedding. In *Proceedings of the 2014 conference on empirical methods in natural language processing (EMNLP)*, pages 1591-1601.

[25] Matthew E Peters, Neumann Mark, et al. 2018. Deep contextualized word representations. arXiv preprint arXiv:1802.05365.